\documentclass[conference]{IEEEtran}
\IEEEoverridecommandlockouts
\usepackage{cite}
\usepackage{amsmath,amssymb,amsfonts}
\usepackage{algorithmic}
\usepackage{graphicx}
\usepackage{textcomp}
\usepackage{xcolor,colortbl}
\usepackage{multirow}
\usepackage{booktabs}
\usepackage{hyperref}

\def\BibTeX{{\rm B\kern-.05em{\sc i\kern-.025em b}\kern-.08em
    T\kern-.1667em\lower.7ex\hbox{E}\kern-.125emX}}

\definecolor{mydarkgray}{rgb}{0.66,0.66,0.66}
\definecolor{mydarkgreen}{rgb}{0.05,0.85,0.05}
\definecolor{mycolor}{rgb}{0.85,0.95,0.85}
\begin{document}

\title{Omni-AD: Learning to Reconstruct Global and Local Features for Multi-class Anomaly Detection}

\author{Jiajie Quan$\textsuperscript{1}$, Ao Tong$\textsuperscript{2}$, Yuxuan Cai$\textsuperscript{2}$, Xinwei He$\textsuperscript{1}$$^\dagger$\thanks{$^\dagger$Corresponding author (Email: \texttt{xwhe@mail.hzau.edu.cn})}, Yulong Wang$\textsuperscript{1}$, Yang Zhou$\textsuperscript{3}$\\

\textsuperscript{1}Huazhong Agricultural University\\
\textsuperscript{2}Huazhong University of Science and Technology\\
\textsuperscript{3}Shenzhen University\\
}

\maketitle


\begin{abstract}
In multi-class unsupervised anomaly detection (MUAD), reconstruction-based methods learn to map input images to normal patterns to identify anomalous pixels. However, this strategy easily falls into the well-known ``learning shortcut'' issue when decoders fail to capture normal patterns and reconstruct both normal and abnormal samples naively. 
To address that, we propose to learn the input features in global and local manners, forcing the network to memorize the normal patterns more comprehensively.
Specifically, we design a two-branch decoder block, named Omni-block. One branch corresponds to global feature learning, where we 
serialize two self-attention blocks but replace the query and (key, value) with learnable tokens, respectively, thus capturing global features of normal patterns concisely and thoroughly. The local branch comprises depth-separable convolutions, whose locality enables effective and efficient learning of local features for normal patterns. 
By stacking Omni-blocks, we build a framework, Omni-AD, to learn normal patterns of different granularity and reconstruct them progressively. 
Comprehensive experiments on public anomaly detection benchmarks show that our method outperforms state-of-the-art approaches in MUAD. Code is available at \url{https://github.com/easyoo/Omni-AD.git}
\end{abstract}


\section{Introduction}
\label{sec:intro}

The past decade has witnessed the rapid development of smart manufacturing. 
Within this realm, Visual Anomaly Detection (VAD) has emerged as an essential component.
Given the difficulty in collecting and labeling anomaly data, previous work operates mainly in an unsupervised manner, such as Embedding-based~\cite{roth2022towards}, Synthesizing-based~\cite{zavrtanik2021draem}, and Reconstruction-based~\cite{you2022unified}.
However, these works are mostly in a single-class setting~\cite{liu2023simplenet}, where a separate VAD model needs to be trained using only normal training images for each product category. 
More recently, multi-class unsupervised anomaly detection (MUAD) has been introduced and has drawn researchers' great attention~\cite{you2022unified, he2024mambaad}. It only needs to train a unified model with normal images of $N$ categories at once, making it more practical and efficient. 

Existing MUAD research mainly focuses on reconstruction strategy. They primarily use a pre-trained backbone to extract features and then train a unified decoder 
to reconstruct the input with learned \emph{normal} patterns. Anomalous pixels can be identified by comparing the input and output. However, to map any input to its normal counterparts, a model has to comprehensively encapsulate both local and global features of diverse normal samples across distinct categories.
Take MVTec-AD~\cite{bergmann2019mvtec} as an example: the decoder is expected to learn \emph{local} repetitive normal patterns in texture classes but also understand the \emph{global} structural integrity of object classes.
This poses great challenges to existing methods, as decoders can learn to take ``shortcuts'' in training.
One way is to build the decoder using the attention mechanism~\cite{vaswani2017attention} to capture long-range dependency. Yet, it is still at the risk of learning ``shortcuts'', as the query, key and value from self-attention are coupled with input.
UniAD~\cite{you2022unified} replaces the query with learnable tokens, forcing them to learn normal pattern distributions. RLR~\cite{he2025learning} choose to replace key and value with learnable ones. 
Yet they all lack sufficient capacity to learn local normal features.
As we know, convolutions are good at extracting fine-grained local spatial features.
\emph{Could we build a decoder with self-attention and convolution, to comprehensively learn both local and global normal patterns?}


Motivated by this, we design a two-branch decoding block named Omni-block with one global branch implemented by modified self-attention with learnable tokens and the other local branch implemented by convolution. 
The global branch aims to fully decouple the query, key, and value from the input with two multi-head attention blocks. To avoid the ``learning shortcut'', we sequentially alternate the decoupling of query and (key, value) with extra learnable tokens. 
The parallel local branch is implemented by depth-wise separable convolutions to effectively and efficiently learn fine-grained local normal features. 
By combining both outputs, we combine the merits of locality in convolution and global relations in MUAD setups. 
Besides, due to the minimal size of learnable tokens in attention, we are able to reduce the computational complexity to be linear with the size of inputs, which makes our network very efficient at handling high-resolution inputs. 
Then, with this dual branch block, we develop a framework named \textbf{Omni-AD} to progressively learn to reconstruct multi-scale normal features, which comprehensively captures both long-range dependencies and different granularities of normal patterns. 
Empirical results on several popular VAD benchmarks show that our framework has obtained remarkable improvements over state-of-the-art methods in the MUAD setting. 

Our contributions are summarized as follows: 
\begin{itemize}
\item We propose a two-branch block named Omni-block, which mitigates ``learning shortcut" issues and efficiently learns both global and local normal features. 
\item We develop Omni-AD upon Omni-blocks to capture different granularities of normal patterns, progressively mapping an input to a normal output. 
\item  Extensive experiments on various popular VAD benchmarks demonstrate the validity and superiority of our design in the MUAD setting. 
\end{itemize}

\section{Related Work}

\noindent\textbf{Unsupervised Anomaly Detection}. Mainstream anomaly detection methods can be broadly categorized into three types: \emph{embedding-based}, \emph{synthesizing-based}, and \emph{reconstruction-based} approaches. 
1) \textbf{Embedding-based methods}~\cite{roth2022towards} leverage pre-trained models to extract features of normal samples and identify anomalies by statistically analyzing deviations in the embedding space. 
For instance, PatchCore~\cite{roth2022towards} utilizes a memory bank to store normal patch features and measures Mahalanobis distance between test sample features and normal features from the memory bank. FastFlow~\cite{yu2021fastflow} employs a 2D flow model to estimate probability distributions and detect anomalies, preserving the spatial relationships of features. 
PaDim~\cite{defard2021padim} utilizes multivariate Gaussian distributions to model the normal embedding space.
However, these works typically require heavy resources and are unfriendly for real-time applications. 
2) \textbf{Synthesizing-based methods}~\cite{zavrtanik2021draem} focus on simulating anomalous regions to generate pseudo-supervisory masks, which helps the model learn to differentiate normal and abnormal distributions. 
For example, DRAEM~\cite{zavrtanik2021draem} generates anomalous regions by embedding diverse masks from other images into normal samples. 
CutPaste~\cite{li2021cutpaste} cuts an image patch and pasts it to another image for training. 
DAF~\cite{cai2023discrepancy} further improves the robustness of existing works by taking discrepancy-aware maps for synthesizing-based methods.
However, these methods have a heavy reliance on the quality of synthesized schemes. Besides, anomalies are unknown, and synthesizing all types of anomalies is impossible.
3) \textbf{Reconstruction-based methods} aim to restore anomalous pixels or features to their corresponding normal representations.  
Models such as  Autoencoder~\cite{ristea2022self}, Generative Adversarial Networks (GANs)~\cite{yan2021learning}, diffusion models~\cite{wyatt2022anoddpm} and Transformer~\cite{mishra2021vt} have been widely utilized. 
Based on the generation quality, anomaly maps can be easily derived by comparing the input and the reconstructed output. 
However, it has been shown that these methods face the ``learning shortcuts'' and sometimes well-restore the anomalies, failing to spot the anomalies by comparisons.
Our method also belongs to this category, but we attempt to alleviate this issue by designing a suitable hybrid decoder with self-attention and depth convolution, forcing the model to focus on comprehensively learning both local and global normal patterns instead of shortcuts. 

\noindent{\textbf{Multi-class Anomaly Detection}}. The seminal work UniAD~\cite{you2022unified} presents the learnable query to alleviate the shortcut issues in reconstruction-based methods. 
OmniAL~\cite{zhao2023omnial} trains a unified model using panel-guided synthetic anomaly data instead of relying solely on normal data. Similarly, HVQ-Trans~\cite{lu2023hierarchical} employs a hierarchical codebook mechanism to mitigate shortcut learning. Additionally, diffusion model demonstrates strong performance~\cite{he2024diffusion}. RLR~\cite{he2025learning} proposes to decouple the key and value from the input. 
In contrast, we incorporate attention with learnable tokens to \emph{fully} detach the query, key, and value. Besides, we also incorporate convolution to enhance the learning of local normal patterns, ensuring a more accurate multi-class anomaly detection.

\section{Method}

\subsection{Network Architecture}
\begin{figure*}
\centering
\includegraphics[width=0.95\linewidth]{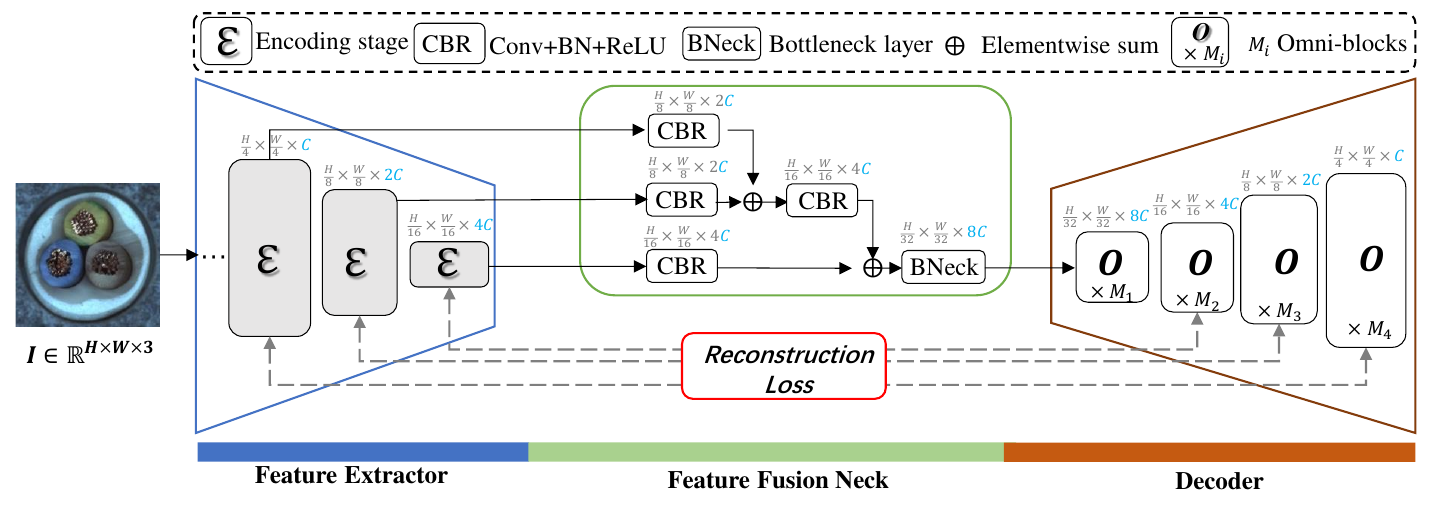}
\caption{\textbf{Overview of Omni-AD. Given an input image, we first use a pretrained network to extract its multi-scale features. Then, we fuse them with the feature fusion neck.  Finally, we feed it into the decoder comprising a series of Omni-blocks to reconstruct multi-scale features progressively.} 
}
\label{fig:arch_overview}
\end{figure*}
Figure~\ref{fig:arch_overview} gives an overview of the proposed Omni-AD. 
As shown, it has three main components: a feature extractor, a feature fusion neck, and a decoder.
We elaborate each below. 

\noindent\textbf{Feature Extractor}. Given an input image $I \in \mathbb{R}^{H \times W \times 3}$, we first feed it into a pretrained CNN $\Phi(\cdot)$ to extract its multi-scale features. Any off-the-shelf CNN pretrained on large-scale datasets like ImageNet~\cite{deng2009imagenet} can be utilized as the feature extractor, such as ResNet~\cite{he2016deep} and EfficientNet~\cite{tan2019efficientnet}. 
In this paper, we follow~\cite{he2024mambaad} and utilize ResNet34~\cite{he2016deep}, which has several feature encoding stages to derive feature maps of decreasing resolution. 
Assuming that $\mathcal{L}$ contains the selected hierarchical stage index subset of $\Phi(\cdot)$ for use, we can denote feature maps of level $l \in \mathcal{L}$ as $\Phi^l \sim \Phi(I_i)^l \in \mathbb{R}^{H_l \times W_l \times C_l}$, where $H_l$, $W_l$ and $C_l$ represent are the height, width, and channel dimensions of the feature maps. In our framework, following~\cite{he2024mambaad}, we only use outputs of the last three stages, denoted as $\mathcal{L}=\{\Phi^{l_k}\}_{k=1}^3$, with $\Phi^{l_k} \in \mathbb{R}^{\frac{H}{2^{k+1}}\times \frac{W}{2^{k+1}} \times 2^{k-1}C}$. 

\noindent\textbf{Feature Fusion Neck}. We aggregate $\{\Phi^{l_k}\}_{k=1}^3$, the feature maps at different hierarchical levels, by feeding them to an H-FPN-style feature fusion neck~\cite{ghiasi2019fpn}. 
The neck consists of several CBR blocks (Conv+BN+ReLU) and one bottleneck layer for progressive fusion.
As shown, given $\{\Phi^{l_k}\}_{k=1}^{3}$, it first integrate $\Phi^{l_1}$ with $\Phi^{l_2}$. Next, the fused features are combined with $\Phi^{l_3}$. Finally, the bottleneck layer is utilized to double the channel dimensions while halving the spatial ones, giving combined features $\Phi^{\text{final}} \in \mathbb{R}^{\frac{H}{32}\times \frac{W}{32} \times 8C}$.
Since $\{\Phi^{l_k}\}_{k=1}^{3}$ have different resolutions, each pixel location in the resulting $\Phi^{\text{final}}$ collects different scales of contextual information. 

 \noindent\textbf{Decoder.} The fused feature maps $\Phi^{\text{final}}$ are further fed into the decoder, which is composed of four stages, with each stage $i$ comprising $M_i$ Omni-blocks (see Sec.~\ref{sec:Omni}). Between these stages, we incorporate upsampling layers to progressively restore the multi-scale spatial resolutions for reconstruction. We adopt feature maps at the last three stages, denoted by $\{\widehat{\Phi}^{l_k}\}_{k=1}^3$.
Finally, we train it with mean squared error (MSE) by summing up reconstruction errors across the three scales. 

\subsection{Omni-block}\label{sec:Omni}

\begin{figure}
\centering
\includegraphics[width=0.95\linewidth]{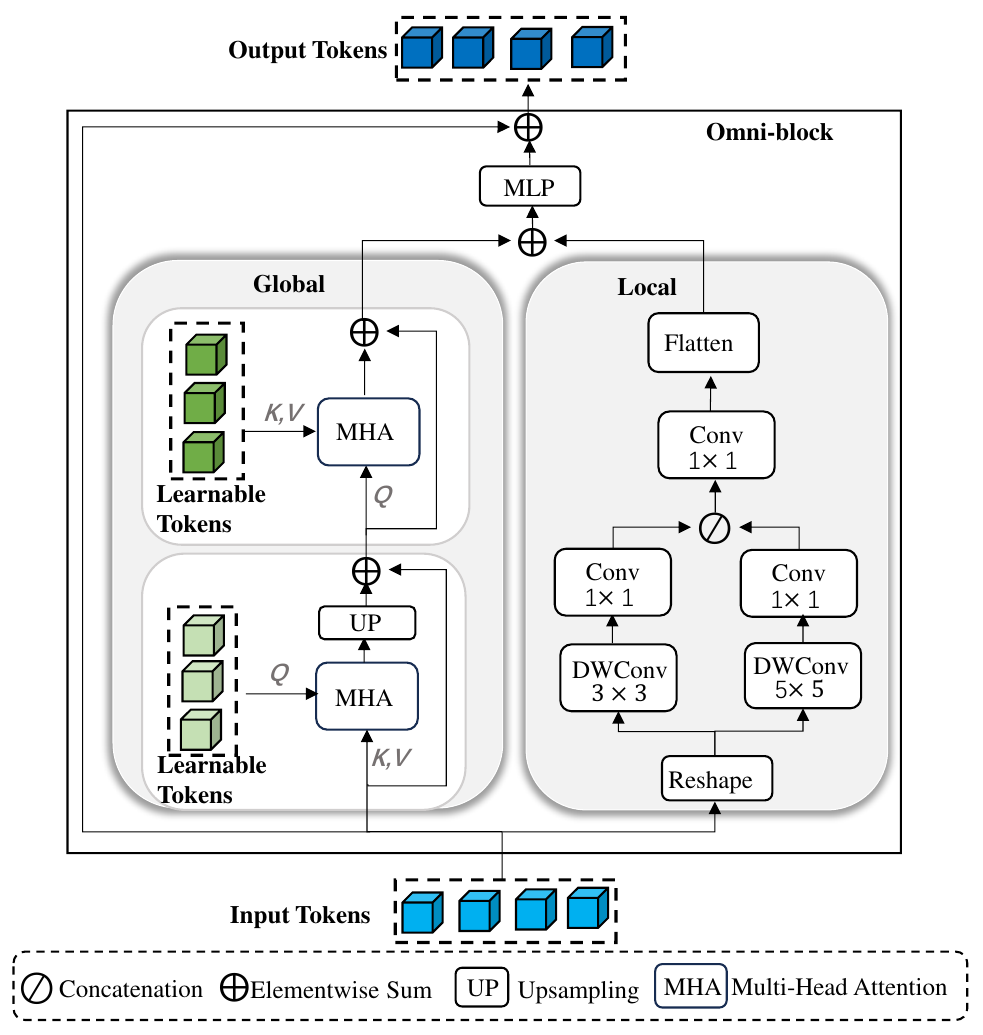}
\caption{\textbf{Detailed structure of the proposed Omni-block.} 
}
\label{fig:block}
\end{figure}

As illustrated in Figure~\ref{fig:block}, our Omni-block is a two-branch module. One branch is designed to learn global relations through two multi-head attention modules. In this branch, learnable tokens are utilized to replace the query ($Q$), key ($K$) and value ($V$) alternatively, making it efficiently learn global normal features while avoiding the ``learning shortcut'' issue.  The other branch learns local features with convolutions, capturing fine-grained normal patterns effectively.
Let $\mathbf{X} \in \mathbb{R}^{N\times D}$ denote the input features, where $N$ is the input token number and $D$ is the token dimension. 
The computing process of Omni-block is described as follows. 

\noindent\textbf{Global branch with learnable tokens}. First, we feed $\mathbf{X}$ to the global branch, which consists of two multi-head attention blocks (MHA). Each MHA~\cite{vaswani2017attention} is just a plain self-attention module, taking Query $Q$, Key $K$, Value $V$ as input:
\begin{equation}
\begin{aligned}
    \mathrm{MHA}(Q, K, V) &= \mathrm{Concat}(\mathrm{head_1}, ..., \mathrm{head_H})W^O\\
    \text{where}~\mathrm{head_i} &= 
    \mathrm{softmax}(\frac{QW_i^{Q}(KW_i^K)^\mathtt{T}}{\sqrt{D_h}})VW^V_i
    \label{eq:mha_equation}
\end{aligned}
\end{equation}
where $H$ is head number, and $i$ is the index, $W_i^Q, W_i^K, W_i^V \in \mathbb{R}^{D \times D_h}$ are weight matrices to project $Q$, $K$ and $V$ for each head, and $W^O \in \mathbb{R}^{HD_h \times D}$ are parameter matrix with $HD_h=D$. In standard self-attention~\cite{vaswani2017attention}, the ($Q$, $K$, $V$) triplet are typically the input features $\mathbf{X}$, which are tightly-coupled to the output and easily leads to the ``learning shortcuts'' issue~\cite{you2022unified}. 
To alleviate this issue, we propose to alternately replace $Q$, ($K$, $V$) with learnable tokens, respectively, for decoupling. 

For the first MHA, we use learnable tokens $\overline{Q} \in \mathbb{R}^{T \times D}$ to harvest context information in the input $\mathbf{X}$, \emph{i.e.}, $\mathbf{F}_1 = \text{MHA}(\overline{Q},\mathbf{X},\mathbf{X}) \in \mathbb{R}^{T \times D}$. Note that $\overline{Q}$ is optimized to capture the normal distributions during training. As a result, it is difficult to reconstruct abnormal samples. 
Another benefit is that $T$ is much smaller than input size $N$, which decreases the key-query dot product complexity from $\mathcal{O}(N^2D)$ to be linear in the input size $\mathcal{O}(NTD)$, and thus is more efficient.   
Finally, we upsampling $\mathbf{F}_1$ with $\mathtt{Up}(\cdot)$ and add a residual connection with input $X$, \emph{i.e.}, $\mathbf{Y}_1 = \mathtt{Up}(\mathbf{F}_1) + \mathbf{X}, \mathbf{Y}_1 \in \mathbb{R}^{N \times D}$.

For the second MHA, we treat $\mathbf{Y}_1$ as the query $Q$ which contains the information from $\mathbf{X}$, while key $K$ and value $V$ coming from shared learnable tokens $\overline{S} \in \mathbb{R}^{T \times D}$, \emph{i.e.}, $K, V$ = $\overline{S}W^K , \overline{S}W^V$, where $W^K, W^V \in \mathbb{R}^{D \times D}$ are projection matrices. 
We finally feed them into MHA to learn normal distributions: $\mathbf{F}_2 = \text{MHA}(\mathbf{Y}_1, K, V)$. Note that the $\overline{S}$ can be regarded as normal reference features during training. At the inference stage, the abnormal tokens deviate from these normal references, which is helpful for anomaly identification. Moreover, it is also efficient due to the linear computational complexity. Similarly, residual connections are incorporated, \emph{i.e.}, $\mathbf{Y}_2 = \mathbf{F}_2 + \mathbf{Y}_1, \mathbf{Y}_2 \in \mathbb{R}^{N \times D}$.

With the above two MHA, we can achieve the decoupling of Query, Key, and Value completely, thus effectively helping the model to understand normal patterns better. 


\noindent\textbf{Local branch with depth-separable convolution.} Complementary to the global branch, a parallel local branch is utilized to enhance the locality information extraction. To this end, we simply adopt depth-wise convolution for the parameter and computation efficiency.    
Specifically, we first reshape $\mathbf{X}$ to restore its spatial size $(h, w)$: $\mathbf{X}' = \mathtt{Reshape}(\mathbf{X}) \in \mathbb{R}^{h \times w \times D}$, where $N = hw$. 
After that, we use Depthwise Separable Convolution, which factorizes standard $k\times k$ convolution into $k\times k$ depth convolution $\mathtt{DWConv}_{k\times k}$ and $1\times1$ point-wise convolution $\mathtt{Conv}_{1\times1}$. We utilize two kernel sizes in parallel: $3\times3
$ and $5\times5$, and then combine the outputs with concatenation after padding and $1\times 1$ convolution for channel reduction. Finally, we flatten the spatial dimensions. The process is formulated below: 
\begin{equation}
  \mathbf{F}_{3} = \mathtt{Conv}_{1\times1} ( \mathtt{DWConv}_{3\times3}(\mathbf{X}^{'}) )
\end{equation}
\begin{equation}
  \mathbf{F}_{4} = \mathtt{Conv}_{1\times1} ( \mathtt{DWConv}_{5\times5}(\mathbf{X}^{'}) )
\end{equation}
\begin{equation}
  \mathbf{F}_{5} = \mathtt{Flatten} (\mathtt{Conv}_{1\times1} (\mathtt{Cat}(\mathbf{F}_{3}, \mathbf{F}_{4} )))
\end{equation}
The local branch facilitates the learning of local patterns. 

Finally, we combine output from local and global branches. 
\begin{equation}
  \mathbf{O}_{\text{final}} = \mathrm{MLP}(\mathbf{Y}_{2} + \mathbf{F}_{5} ) + \mathbf{X}
\end{equation}

\subsection{Training and Inference}
For training, we utilize Mean Square Error to measure the loss between reconstructed features  $\{\widehat{\Phi}^{l_k}\}_{k=1}^3$ and the original extracted features $\{\Phi^{l_k}\}_{k=1}^3$. The loss function is defined as
\begin{equation}
  \mathcal{L} = \sum_{k=1}^{3} \frac{\lVert \Phi^{l_k} - \widehat{\Phi}^{l_k} \rVert _2^2}{H_k W_k},
  \label{eq:loss}
\end{equation}

For inference, we combine cosine similarities across multiple scales, specifically at 2, 3, and 4 stages, to derive the anomaly maps.

\section{Experiments}

\subsection{Experimental Setup} 

\noindent\textbf{Datasets.}
\textbf{MVTec-AD}~\cite{bergmann2019mvtec} consists of 15 categories, including 5 texture types and 10 object types. It has 3,629 normal images for training, with 467 normal and 1,258 anomalous images for testing. \textbf{VisA}~\cite{zou2022spot} has a total of 10,821 images divided into 12 objects, including 9,621 normal samples and 1,200 anomalous samples. 
\textbf{Real-IAD}~\cite{wang2024real} is largest industrial anomaly detection dataset. It comprises 30 distinct objects, with 36,465 normal images for training and 114,585 images for testing, including 63,256 normal and 51,329 anomalous samples.

\noindent\textbf{Metric.} Following~\cite{zhang2024ader,he2024mambaad}, we report Area Under the Receiver Operating Curve (AU-ROC)~\cite{zavrtanik2021draem}, Average Precision (AP)~\cite{zavrtanik2021draem}, and F1-score (F1\_max)~\cite{zou2022spot}, at the image level. For pixel-level, we also report Area Under the Per-Region-Overlap (AU-PRO)~\cite{bergmann2020uninformed}.
We calculate the average anomaly detection score (mAD)~\cite{zhang2023exploring} to assess overall performance.

\noindent\textbf{Implementation Details.} 
All input images are resized to 256$\times$256. Following MambaAD~\cite{he2024mambaad}, we use a pre-trained ResNet-34~\cite{he2016deep} for feature extraction. The decoder consists of four stages, with 3, 9, 9, and 7 Omni blocks stacked in each stage, respectively. Stages 2-4 start with spatial upsampling. We use AdamW with learning rate $1\times10^{-3}$ and weight decay $1\times10^{-4}$ and train for 500 epochs. 

\subsection{Main Results}
\begin{table*}[ht]
	\centering
        \small 
	\caption{Performance comparisons (\%) with state-of-the-arts on different AD datasets for multi-class setting.}
	\resizebox{\linewidth}{!}{
		\begin{tabular}{ccccccccccc}
			\toprule
			\multirow{2}[2]{*}{Dateset} & \multirow{2}[2]{*}{Method} &  \multirow{2}[2]{*}{Source} & \multicolumn{3}{c}{Image-level} & \multicolumn{4}{c}{Pixel-level}& \multirow{2}[2]{*}{\textbf{mAD}} \\
			\cmidrule(r){4-6} \cmidrule(l){7-10} 
			& &&\multicolumn{1}{c}{AU-ROC} & \multicolumn{1}{c}{AP} & \multicolumn{1}{c}{F1\_max} & \multicolumn{1}{c}{AU-ROC} & \multicolumn{1}{c}{AP} & \multicolumn{1}{c}{F1\_max} & \multicolumn{1}{c}{AU-PRO} \\
			\hline
			\multirow{8}[0]{*}{MVTec-AD~\cite{bergmann2019mvtec}} & RD4AD~\cite{deng2022anomaly} & CVPR'22 & 94.6  & 96.5  & 95.2  & 96.1  & 48.6  & 53.8  & 91.1 &82.3\\
			& UniAD~\cite{you2022unified} & NeurlPS'22 & 96.5  & 98.8  & 96.2  & 96.8  & 43.4  & 49.5  & 90.7& 81.7\\
			& SimpleNet~\cite{liu2023simplenet} & CVPR'23  & 95.3  & 98.4  & 95.8  & 96.9  & 45.9  & 49.7  & 86.5& 81.2\\
			& DeSTSeg~\cite{zhang2023destseg} & CVPR'23 & 89.2  & 95.5  & 91.6  & 93.1  & 54.3  & 50.9  & 64.8& 77.1\\
			& DiAD~\cite{he2024diffusion}  & AAAI'24 & 97.2  & 99.0  & 96.5  & 96.8  & 52.6  & 55.5  & 90.7& 84.0\\
            & RLR~\cite{he2025learning} & ECCV'24  & \underline{98.6} & - & - & \textbf{98.5} & - & - & - & -
             \\
			& MambaAD~\cite{he2024mambaad} & NeurlPS'24  & \underline{98.6} & \underline{99.6} & \underline{97.8} & 97.7 & \underline{56.3} & \underline{59.2} & \underline{93.1}& \underline{86.0}\\
			& \cellcolor{mycolor} Omni-AD & \cellcolor{mycolor} -  & \cellcolor{mycolor} \textbf{99.0} & \cellcolor{mycolor} \textbf{99.7} & \cellcolor{mycolor} \textbf{98.3} & \cellcolor{mycolor} \underline{97.9} & \cellcolor{mycolor} \textbf{56.8} & \cellcolor{mycolor} \textbf{59.9} & \cellcolor{mycolor} \textbf{93.4}& \cellcolor{mycolor} \textbf{86.4}\\
			\hline
			\multirow{7}[0]{*}{VisA~\cite{zou2022spot}} & RD4AD~\cite{deng2022anomaly} & CVPR'22 & 92.4  & 92.4  & \underline{89.6} & 98.1 & 38.0  & 42.6  & \underline{91.8}&77.8 \\
			& UniAD~\cite{you2022unified} & NeurlPS'22 & 88.8  & 90.8  & 85.8  & 98.3  & 33.7  & 39.0  & 85.5& 74.6\\
			& SimpleNet~\cite{liu2023simplenet} & CVPR'23 & 87.2  & 87.0  & 81.8  & 96.8  & 34.7  & 37.8  & 81.4&72.4 \\
			& DeSTSeg~\cite{zhang2023destseg} & CVPR'23 & 88.9  & 89.0  & 85.2  & 96.1  & \textbf{39.6} & \underline{43.4}  & 67.4&72.8 \\
			& DiAD~\cite{he2024diffusion}  & AAAI'24 & 86.8  & 88.3  & 85.1  & 96.0  & 26.1  & 33.0  & 75.2& 70.1\\
			& MambaAD~\cite{he2024mambaad} & NeurlPS'24 & \underline{94.3} & \underline{94.5} & 89.4  & \underline{98.5} & \underline{39.4}  & \textbf{44.0} & 91.0 &\underline{78.7}\\
			& \cellcolor{mycolor} Omni-AD  &\cellcolor{mycolor} - & \cellcolor{mycolor} \textbf{94.6} & \cellcolor{mycolor} \textbf{95.0} & \cellcolor{mycolor} \textbf{90.3} & \cellcolor{mycolor} \textbf{98.8} & \cellcolor{mycolor} {38.7} & \cellcolor{mycolor} 43.3 & \cellcolor{mycolor} \textbf{91.9}& \cellcolor{mycolor} \textbf{78.9}\\
			\hline
			\multirow{6}[0]{*}{Real-IAD~\cite{wang2024real}}& 
			 UniAD~\cite{you2022unified}& NeurlPS'22 & 83.0  & 80.9  & 74.3  & 97.3  & 21.1  & 29.2  & 86.7 &67.5\\
			& SimpleNet~\cite{liu2023simplenet}& CVPR'23 &57.2  & 53.4  & 61.5  & 75.7  & \;\;2.8   & \;\;6.5   & 39.0 &42.3\\
			& DeSTSeg~\cite{zhang2023destseg}& CVPR'23 &82.3  & 79.2  & 73.2  & 94.6  & \textbf{37.9} & \textbf{41.7} & 40.6 &64.2\\
		& DiAD~\cite{he2024diffusion}& AAAI'24 &75.6  & 66.4  & 69.9  & 88.0  &  \;\;2.9   &  \;\;7.1   & 58.1 &52.6\\
		& MambaAD~\cite{he2024mambaad} & NeurlPS'24 &\underline{86.3} & \underline{84.6} & \underline{77.0} & \underline{98.5} & 33.0  & 38.7  & \underline{90.5} &\underline{72.7}\\
			& \cellcolor{mycolor} Omni-AD  & \cellcolor{mycolor} - & \cellcolor{mycolor} \textbf{88.2} & \cellcolor{mycolor} \textbf{86.5} & \cellcolor{mycolor} \textbf{79.1} & \cellcolor{mycolor} \textbf{98.8} & \cellcolor{mycolor} \underline{34.3} & \cellcolor{mycolor} \underline{39.7} & \cellcolor{mycolor} \textbf{91.6}& \cellcolor{mycolor} \textbf{74.0}\\
			\bottomrule
		\end{tabular}%
	}
	\label{tab:main_results}%
    \vspace{-12pt}
\end{table*}%
We compare against MUAD methods, \emph{i.e.}, UniAD~\cite{you2022unified}, DiAD~\cite{he2024diffusion}, RLR~\cite{he2025learning} and MambaAD~\cite{he2024mambaad}. We also compare with other superior reconstruction-based method RD4AD~\cite{deng2022anomaly} and embedding-based ones (DeSTseg~\cite{zhang2023destseg} and SimpleNet~\cite{liu2023simplenet}). 

As shown in Table~\ref{tab:main_results}, on \textbf{MVTec-AD}, our approach outperforms all the compared methods. At the image level, it obtains detection performance of 99.0/99.7/98.3, and at the pixel level, it attains segmentation of 97.9/56.8/59.9/93.4.
Specifically, compared with previous state-of-the-art method MambaAD, we achieve increases of 0.4/0.1/0.5 in detection and 0.2/0.5/0.7/0.3 in segmentation. The overall mAD metric improves by 0.4 and 2.4 when compared with MambaAD and DiAD, respectively. 
On the challenging \textbf{VISA} dataset, our method also attains the leading performance, where all our metrics significantly exceed DiAD, with mAD improving by 8.8. It also has a better average performance than MambaAD, improving mAD by 0.2. 
Finally, on the largest \textbf{Real-IAD} dataset, we remarkably surpass the best previous method MambaAD by 1.3\% mAD.
These results consistently demonstrate the superiority of the proposed Omni-AD.

\subsection{Ablation Studies}

This section analyzes the component effectiveness in our design.
All experiments are done on the MVTec-AD dataset. 

 \begin{table}[h]
    \centering
    \caption{Ablation Study of Learnable Q and KV.}
    \resizebox{\linewidth}{!}{
    \begin{tabular}{cccccccccc}
       \toprule
     \multirow{2}[2]{*}{Q} & \multirow{2}[2]{*}{KV} & \multicolumn{3}{c}{Image-level} & \multicolumn{4}{c}{Pixel-level} \\
     \cmidrule(lr){3-5} \cmidrule(lr){6-9}  & & AU-ROC & AP & F1\_max & AU-ROC & AP & F1\_max & AU-PRO 
     \\ \hline
    - & - & 98.4 & 99.4 & 97.8 & 97.6 & 54.0 & 57.6 & 91.9 \\ 
    $\checkmark$ & ~ & 98.6 & 99.5 & 98.2 & \textbf{98.0} & 56.6 & 59.7 & 93.1 \\ 
   $\checkmark$ & $\checkmark$ & \textbf{99.0} & \textbf{99.7} & \textbf{98.3} & 97.9 & \textbf{56.8} & \textbf{59.9} & \textbf{93.4} \\ 
    \hline
    \end{tabular}
}
\label{tab:ablation_learnable_q_kv}%
\end{table}
\noindent\textbf{Effect of learnable tokens.} Table~\ref{tab:ablation_learnable_q_kv} studies the role of learnable tokens. As shown, when learnable tokens are absent (row-1), \emph{i.e.}, using plain self-attention, the performance only reaches 54.0 AP at pixel-level. When replacing the query with learnable tokens, we increase this metric to 56.6 (+2.6\%). All the other metrics are also improved. Further replacing key and value with learnable keys (row 3) can consistently augment both anomaly detection and localization on 6/7 metrics, with comparable AU-ROC at pixel level, demonstrating the effectiveness of replacing both Q and KV with learnable tokens. 

\noindent\textbf{Effect of learnable tokens position.} Table~\ref{tab:ablation_combinations} studies the impact of replacing Q and KV with learnable tokens in various orders, denoted as A+B for replacements. While these changes have a minor impact on detection performance, \emph{they significantly influence localization performance}. The Q+KV combination achieves the best results in both aspects across 6 metrics, making it as the default configuration.
\begin{table}[h]
    \centering
    \caption{Ablation Study of the Combination of Learnable Q and K,V.}
    \resizebox{\linewidth}{!}{
    \begin{tabular}{ccccccccc}
       \toprule
        \multirow{2}[2]{*}{Combination} & \multicolumn{3}{c}{Image-level} & \multicolumn{4}{c}{Pixel-level} \\
     \cmidrule(r){2-4} \cmidrule(l){5-8} 
        & AU-ROC & AP & F1\_max & AU-ROC & AP & F1\_max & AU-PRO 
        \\ \hline
        Q~+~Q & \textbf{99.0} & \textbf{99.7} & \textbf{98.4} & \textbf{98.0} & 55.0 & 59.0 & 92.9 \\ 
        KV~+~KV & 98.9 & 99.6 & \textbf{98.4} & 97.7 & 51.2 & 56.6 & 91.4 \\ 
        KV~+~Q & \textbf{99.0} & \textbf{99.7} & \textbf{98.4} & \textbf{98.0} & 55.4 & 59.1 & 93.0 \\ 
        Q~+~KV & \textbf{99.0} & \textbf{99.7} & 98.3 & 97.9 & \textbf{56.8} & \textbf{59.9} & \textbf{93.4} \\ \hline 
    \end{tabular}
}
\label{tab:ablation_combinations}%
\vspace{-12pt}
\end{table}

\noindent\textbf{Effect of learnable token numbers.} 
Table~\ref{tab:ablation_number} shows our method is robust to the number of learnable tokens. Setting it to 64 achieves the best performance with comparable complexity, making it the default choice.
\begin{table}[ht]
    \centering
       \setlength{\tabcolsep}{12pt}
    \caption{Ablation Study of the Number of Learnable Tokens.}
    \resizebox{0.85\linewidth}{!}{
    \begin{tabular}{cccc}
       \toprule
     Number &Params(M)&FLOPs(G)& 
     mAD\\ \hline
     16 &29.4 &10.1 & 
     86.2 \\ 
     36 &29.6 &10.5 & 
     86.2 \\ 
     64 &29.9 & 11.2& 
     86.4 \\  
     100 &30.3 &12.0 & 
     86.2 \\  
     \hline
    \end{tabular}
}
\label{tab:ablation_number}%
\end{table}
\begin{table}[ht]
    \centering
    \caption{Ablation Study of Global and Local Branches.}
    \resizebox{\linewidth}{!}{
    \begin{tabular}{cccccccccc}
       \toprule
     \multirow{2}[2]{*}{Global} & \multirow{2}[2]{*}{Local} & \multicolumn{3}{c}{Image-level} & \multicolumn{4}{c}{Pixel-level} \\ 
     \cmidrule(r){3-5} \cmidrule(l){6-9}  & & AU-ROC & AP & F1\_max & AU-ROC & AP & F1\_max & AU-PRO \\ 
     \hline
    \checkmark & $\times$ & 98.1 & 99.3 & 97.5 & 97.8 & 56.1 & 59.1 & 93.0 \\ 
    $\times$ & \checkmark & 98.8 & 99.6& 98.2 & 97.4 & 49.3 & 55.6 & 90.8 \\ 
    \checkmark & \checkmark & \textbf{99.0} & \textbf{99.7} & \textbf{98.3} & \textbf{97.9} & \textbf{56.8} & \textbf{59.9} & \textbf{93.4} \\ \hline 
    \end{tabular}
}
\label{tab:ablation_branch}
\end{table}

\noindent\textbf{Effect of local and global branch.} As shown in Table~\ref{tab:ablation_branch}, when the local branch is disabled, the model's detection capability decreases, but it maintains good localization ability, reaching 93.0\% AU-PRO. When the global branch is disabled, the model still retains strong detection performance, but its localization ability significantly decreases, leading to an AU-PRO reduction of 2.6\%. This indicates that the global branch plays a crucial role in anomaly localization, while the local branch significantly enhances detection performance. 
\begin{table}[!htb]
\small 
    \centering
     \setlength{\tabcolsep}{12pt}
    \caption{Ablation Study of Decoder Depth at Each Stage.}
    \resizebox{0.85\linewidth}{!}{
    \begin{tabular}{cccc}
       \toprule
     Depths  & 
     Params(M)&FLOPs(G) & mAD\\ \hline
     2-2-2-2~(8) & 
     19.0& 6.3 & 85.1 \\ 
     3-6-6-3~(18) & 
     26.5& 8.7 & 86.2 \\ 
     3-9-9-7~(28) & 
     29.9 & 11.2 & 86.4 \\  
     6-12-11-9~(38) & 42.7 & 13.7 & 86.3\\ \hline
    \end{tabular}
}
\label{tab:ablation_depths}%
\end{table}

 \begin{table}[!htb]
 \small 
    \centering
     \setlength{\tabcolsep}{12pt}
    \caption{Ablation Study on ResNet Series Encoders}
    \resizebox{0.85\linewidth}{!}{
    \begin{tabular}{cccc}
       \toprule
     Encoder  &  Params(M)&FLOPs(G) & mAD\\ \hline
     ResNet18 &  24.5 & 9.0 & 85.9 \\ 
     ResNet34 & 29.9 & 11.2 & 86.4 \\ 
     ResNet50 &  468 & 114 & 85.5 \\  
     WideResNet50 & 484 & 121 & 86.3\\ \hline
    \end{tabular}
}
\label{tab:ablation_encoder}%
\vspace{-12pt}
\end{table}

\begin{figure}[ht]
    \centering
\includegraphics[width=0.95\linewidth]{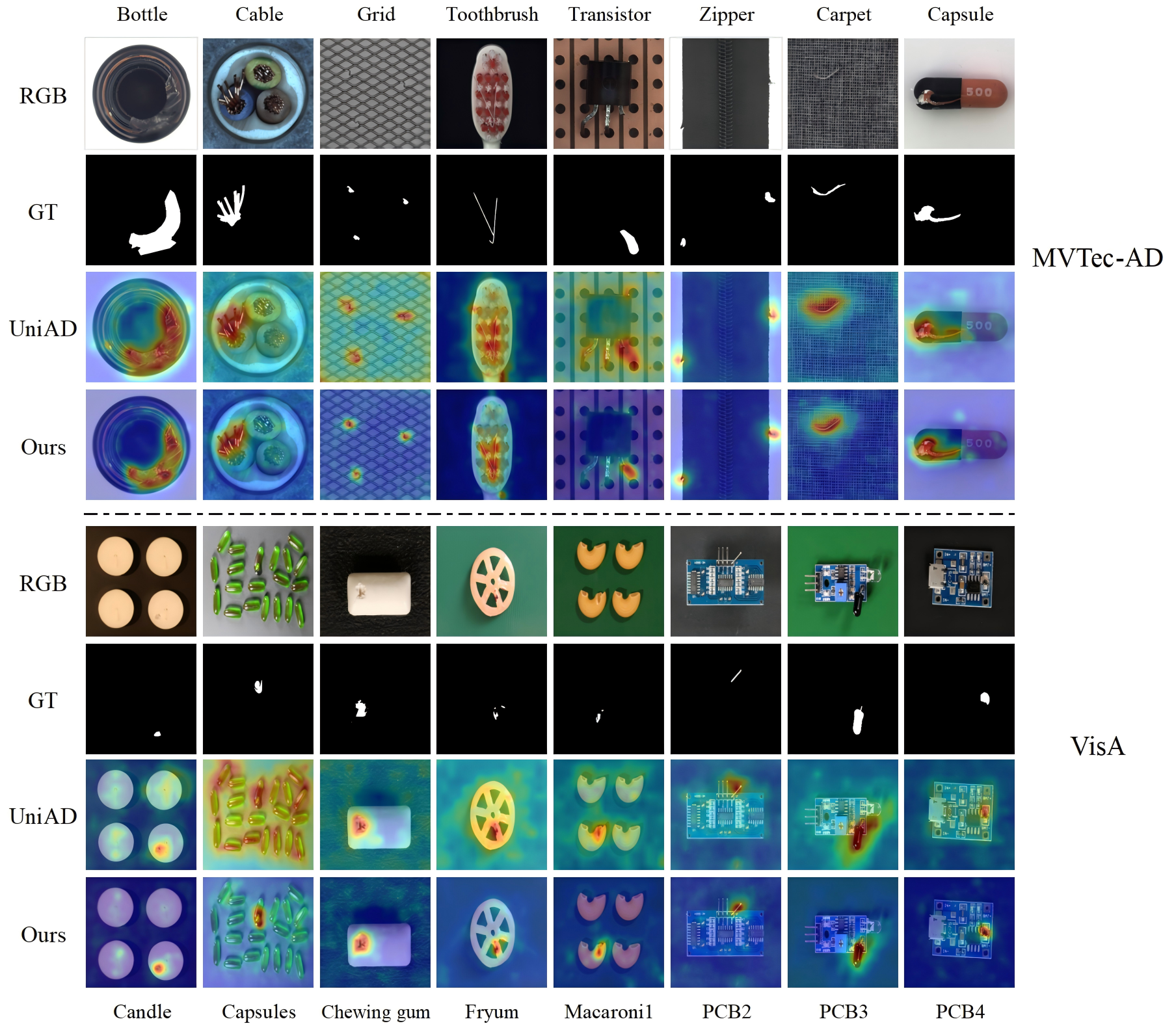}
    \caption{Qualitative results on MVTec and VisA.}
    \label{fig:visual}
    \vspace{-12pt}
\end{figure} 

\noindent\textbf{Effect of decoder depth at each stage.} Table~\ref{tab:ablation_depths} shows that increasing the decoder depth improves performance but also raises computational complexity and parameter overhead. A depth of 3-9-9-7 offers the best results with a manageable increase in complexity, making it our default configuration.

\noindent\textbf{Effect of backbones.} 
Table~\ref{tab:ablation_encoder} shows that using ResNet34 as the backbone achieves the highest performance and remarkable efficiency compared to others. ResNet18 is an excellent choice for prioritizing efficiency.

\noindent\textbf{Visualization}.
Figure~\ref{fig:visual} visualize results on MVTec-AD and VisA. Our framework successfully detects the visual anomalies of diverse structures and appearances across various objects more precisely, demonstrating its strong capacity.

\section{Conclusion And Future Work}

We propose Omni-AD for multi-class unsupervised anomaly detection, which consists of a pre-trained encoder, a feature fusion neck, and a decoder that learns to reconstruct multi-scale normal features from local and global perspectives. At the core of our decoder, a two-branch block named Omni is introduced. It has a local branch learning normal features of
local granularity with depth-convolution layers. The remaining global branch captures normal patterns with learnable tokens.
Omni-AD has attained outstanding performance on public AD benchmarks. In the future, we plan to extend it to other domains, such as medical imaging, to demonstrate its generality.

\section{Acknowledgment}

This work is supported by Hubei Province Natural Science Foundation (No.2023AFB267); National Natural Science Foundation of China (No.62302188); Fundamental Research Funds for the Central Universities (No.2662023XXQD001).

\bibliographystyle{IEEEbib}
\bibliography{icme2025references}

\end{document}